\documentclass{iet-ell}

\usepackage{url}
\usepackage{graphicx}
\usepackage{amsfonts}
\usepackage{amsmath}
\usepackage{algorithm,algpseudocode}
\usepackage[caption=false,font=tiny,labelfont=sf,textfont=sf]{subfig}

\usepackage{nameref} % prints the section title

\setcopyright{open}
\ietVolume{00}
\ietYear{2025}
\ietDoi{ell2.10001}

\received{10 October 2025}
\accepted{1 January 2026}

\begin{document}

\title{Perspective-aware fusion of incomplete depth maps and surface normals for accurate 3D reconstruction}

\author[af1]{Ondrej Hlinka}
\orcid{0009-0007-9550-7750}
\author[af1]{Georg Kaniak}
\orcid{0009-0005-6095-6906}
\author[af1]{Christian Kapeller}
\orcid{0000-0001-6034-8402}

\affil[af1]{Competence unit High-Performance Vision System, Center for Vision, Automation and Control, Austrian Institute of Technology, Giefinggasse 4, 1210 Vienna, Austria}

\corresp{Email:  ondrej.hlinka@ait.ac.at}

\begin{abstract}
We address the problem of reconstructing 3D surfaces from depth and surface normal maps acquired by a sensor system based on a single perspective camera. 
Depth and normal maps can be obtained through techniques such as structured-light scanning and photometric stereo, respectively.
We propose a perspective-aware log-depth fusion approach that extends existing orthographic gradient-based depth-normals fusion methods
by explicitly accounting for perspective projection, leading to metrically accurate 3D reconstructions. Additionally, the method handles missing depth measurements by leveraging available surface normal information to inpaint gaps.
Experiments on the DiLiGenT-MV data set demonstrate the effectiveness of our approach and highlight the importance of perspective-aware depth-normals fusion.
\end{abstract}

\maketitle

\vspace{0mm}
\section{Introduction}

This letter addresses the problem of fusing a depth map (acquired with e.g., a structured-light (SL) \cite{zanuttigh2016operating} scanner) and a surface normal map (obtained with e.g., photometric stereo (PS) \cite{woodham1980photometric,ackermann2015survey}) into a single refined depth map.  
We assume that both maps are acquired by a sensor system based on a single perspective camera (e.g., \cite{haque2014high}) and hence are \mbox{pixel-aligned}.
The fused depth map simultaneously inherits  
the high-frequency geometric detail contained in the PS normals, and  
the absolute depth provided by the SL system.  
As demonstrated in \cite{antensteiner2018review}, such fusion can significantly enhance 3D reconstruction. However, many existing approaches
\cite{antensteiner2017high,antensteiner2018review,antensteiner2018variational}
assume orthographic projection, which can lead to distortions in real-world setups, where perspective effects are not negligible. 

To address this limitation, 
we propose a perspective-aware depth-normals fusion approach that extends any gradient-based orthographic method to perspective projection via a log-depth substitution. A common challenge in practical SL setups is the presence of depth gaps \cite{wang2016filling,niu2025novel,traxler2021experimental,yu2013shading} due to occlusions, reflective or SL-pattern absorbing surfaces. Our approach naturally accommodates such missing or unreliable depth data, enabling depth reconstruction inside gaps solely from surface normals within a unified, perspective-aware framework.

\vspace{0mm}
\subsection{Naive perspective fusion}
\label{sec:alt_fusion}

A straightforward approach to adapt orthographic fusion techniques to perspective projection involves back-projecting the perspective depth map into a 3D point cloud and then reprojecting it onto an orthographic plane, yielding an orthographic depth map. Standard orthographic fusion methods \cite{antensteiner2018review,antensteiner2018variational,antensteiner2017high} can then be applied. However, this simplistic approach presents several limitations: 
i.) reprojecting the 3D points onto an orthographic plane to form a depth map requires interpolation, 
which introduces errors; 
ii.) furthermore, 
where depth is unknown, no 3D points can be reconstructed,
so ad-hoc depth inpainting/interpolation is required before fusion, leading to additional errors; 
iii.) finally, the surface normals need to be interpolated as well, to ensure pixel alignment with the reprojected orthographic depth map. 
Our log-depth fusion formulation
avoids all of these issues.
It naturally handles missing depth information and does not require interpolation or ad-hoc pre-fusion inpainting.

\vspace{0mm}
\subsection{Related work}

\vspace{0.0mm}

In \cite{queau2018normal,cao2022bilateral,kim2024discontinuity}, integration of surface normals under both orthographic and perspective projection is addressed.
These approaches reformulate the normal integration problem in the logarithmic depth domain, enabling the use of orthographic solvers such as \cite{frankot1988method} or \cite{horn1986variational} in perspective projection settings.
However, in contrast to our work, the approaches \cite{queau2018normal,kim2024discontinuity} focus solely on normal integration and do not consider the more general problem of fusing normals with depth data.
The source code \cite{cao2022code} released with \cite{cao2022bilateral} includes an option to incorporate a log-depth prior, thus enabling a basic form of depth-normals fusion. This is achieved in \cite{cao2022code} by augmenting the pure normal integration objective function from \cite{cao2022bilateral} with a quadratic data term based on the log-depth. 
In contrast to directly extending a normal integration objective function, we derive a systematic logarithmic domain approach to depth-normals fusion in a structured and  principled way, and we demonstrate how to apply it to extend any existing orthographic gradient-based fusion method
to handle perspective projection. 
Furthermore, the perspective-aware fusion methods that we obtain by applying this methodology to the orthographic methods from \cite{antensteiner2018review,antensteiner2018variational,antensteiner2017high},  
employ a different objective function in the underlying optimization problem than the fusion extension from \cite{cao2022code}.
Finally, we demonstrate that our perspective-aware fusion methods can achieve an improved performance over the fusion extension from \cite{cao2022code} on the DiLiGenT-MV data set with emulated incomplete depth information.

Recently, neural surface normal integration was introduced in \cite{cao2024supernormal}. However, unlike our approach, it requires multi-view inputs and is computationally intensive, and therefore cannot be easily applied to single-view real-time setups.

Depth and normals fusion under perspective projection has been explored in prior work \cite{nehab2005efficiently,zhang2012edge,haque2014high,yu2019depth}. In \cite{nehab2005efficiently}, a perspective-aware fusion technique is proposed, but the method does not leverage the log-depth substitution. Furthermore, the proposed method does not handle missing depth data. While the extension to handle missing depth data is relatively straight forward, it is not considered explicitly in \cite{nehab2005efficiently}. The methods in \cite{zhang2012edge,haque2014high,yu2019depth} extend the fusion approach from \cite{nehab2005efficiently}, but similarly do not leverage the log-depth substitution and do not address the handling of missing depth data.

\vspace{-1mm}
\subsection{Contribution}

The main contributions of this letter are as follows:
\vspace{-2.5mm}
\begin{itemize}
    \item We introduce a principled log-depth approach to depth and normal fusion. We show that adopting a log-domain formulation is beneficial because it
enables the reuse of any existing gradient-based orthographic solver, such as those presented in \cite{antensteiner2017high,antensteiner2018review,antensteiner2018variational}, in a perspective-aware fusion framework.
    \item Our approach inherently handles missing or unreliable depth data by performing perspective-aware inpainting relying solely on surface normals. This eliminates the need for ad-hoc pre-fusion interpolation or inpainting methods.
    \item To demonstrate improvements over the state-of-the-art methods, we evaluate our approach on the DiLiGenT-MV data set.
\end{itemize}

\vspace{-2mm}

\section{Definitions and notation}

For clarity, we write all maps as continuous functions
and use analytic
derivatives such as
$\partial_u$ and $\partial_v$.
However, in
implementation, the image domain is the integer pixel lattice
\(\mathcal{P}=\{(u,v) \in \mathbb{Z}^2\}\) 
with unit spacing
\(\Delta u=\Delta v=1\) pixel, and all spatial derivatives
are evaluated
with finite differences.  We keep the
continuous notation to avoid clutter, but the reader should bear in mind this
discrete realization throughout the derivations that follow.

Let $\mathbf{s} = [x, y, z]^\top \in \mathbb{R}^3$ be a surface point in 3D space, and let $\mathbf{n}(\mathbf{s}) =
[n_x, n_y, n_z]^\top \in \mathbb{R}^3$ be the unit surface normal vector at the surface point $\mathbf{s}$.
When the surface is observed by a camera, the point $\mathbf{s}$ and its normal vector $\mathbf{n}(\mathbf{s})$
are projected onto a point $\mathbf{u} = [u, v]^\top \in \mathbb{R}^2$ in the image plane.
We can parameterize the surface and its \textbf{normal map} as vector-valued functions of the image plane coordinates $u$ and $v$: \vspace{-1mm}
\begin{equation}
\mathbf{s}\colon\mathbb{R}^2 \rightarrow \mathbb{R}^3, \quad
\mathbf{s}(u,v)= [x(u,\!v),  y(u,\!v), z(u,\!v)]^\top\!\!, \label{eq:3D_point} \vspace{-0.5mm}
\end{equation}
and \vspace{-2mm}
\begin{equation}
\mathbf{n}\colon\mathbb{R}^2 \rightarrow \mathbb{R}^3,\quad
\mathbf{n}(u,v)= [n_x(u,\!v),n_y(u,\!v),n_z(u,\!v)]^\top\!\!,
\label{eq:normal_map}
\end{equation}
respectively.
Furthermore, we define a \textbf{depth map} as the scalar field \vspace{-1mm}
\begin{equation}
d\colon\mathbb{R}^2 \rightarrow \mathbb{R},\qquad d(u,v)\;=\;z(u,v), %\nonumber
\end{equation}
i.e., it stores for every image point $\mathbf{u}$ the z-coordinate of the corresponding surface point  \eqref{eq:3D_point}.
Finally, we define the \linebreak \textbf{depth gradient field} as \vspace{-2mm}
\begin{equation}
\mathbf{g}\colon\mathbb{R}^2 \rightarrow \mathbb{R}^2,\quad
\mathbf g(u,v)\;=\;\nabla d(u,v)
=\!\begin{bmatrix}\partial_u d(u,v)\\\partial_v d(u,v)\end{bmatrix}.
\label{eq:gradient_cont}
\end{equation}

\vspace{-3.5mm}
\section{Orthographic fusion}
\label{sec:ortho_fusion}

If a depth map was captured with an orthographic (parallel-projection) sensor,
the corresponding 3D point \eqref{eq:3D_point} is simply  
\begin{equation}
\mathbf{s}_o(u,v)=
[u, v, d(u,v)]^\top.
\label{eq:orthographic_point} %\nonumber
\end{equation}
Since the surface normal vector must be parallel to the cross product of the surface tangent vectors
$\partial_u\mathbf s_o(u,v) =\![1,0,\partial_u d(u,v)]^\top$ and $\partial_v\mathbf s_o(u,v) =\![0,1,\partial_v d(u,v)]^\top$, we can write \vspace{-3mm}
\begin{equation}
\mathbf{n}_o(u,v) \propto \partial_u\mathbf s_o (u,v) \times \partial_v\mathbf s_o (u,v) = 
\begin{bmatrix}
-\partial_u d(u,v) \\[2pt]
-\partial_v d(u,v) \\[2pt]
 1
\end{bmatrix}.
\label{eq:normal_def}
\end{equation} %\vspace{-1mm}
Scaling \eqref{eq:normal_map} by its third component $n_z(u,\!v)$, we can write $\mathbf{n}_o(u,v) \propto [n_x(u,\!v)/n_z(u,\!v), n_y(u,\!v)/n_z(u,\!v), 1]^\top$. By component-wise comparison with \eqref{eq:normal_def}, we can see that $\partial_u d(u,v) = -n_x(u,\!v)/n_z(u,\!v)$ and $\partial_v d(u,v) = -n_y(u,\!v)/n_z(u,\!v)$.
Hence, we see that under orthographic projection, the relationship between the normal vector and depth map gradient vector in \eqref{eq:gradient_cont} 
is (see also \cite{antensteiner2018review,cao2022bilateral}): \vspace{-2.0mm}
\begin{equation}
\mathbf g_o(u,v)=
\begin{bmatrix}
\partial_{u} d(u,v) \\[2pt] \partial_{v} d(u,v)
\end{bmatrix}
=
-\frac1{n_{z} (u,v)}
\begin{bmatrix}
n_{x} (u,v)\\[2pt]n_{y} (u,v)
\end{bmatrix}.
\label{eq:grad_from_normal}
\end{equation}
Thus \eqref{eq:grad_from_normal} establishes a relationship between the observed normals (obtained e.g., using PS) with the observed depth map (obtained e.g., from SL).

Let $d_\mathrm{obs}(u,v)$ be an observed depth map with an accompanying confidence map $\kappa(u,v)\in[0,1]$; $\kappa(u,v)=0$ indicates missing depth, for instance due to occlusions, specular reflections of the SL pattern, or absorption by non-cooperative materials. 
For SL setups, $\kappa(u,v)>0$ typically indicates the uniqueness (and hence the reliability) of the pattern matching process at pixel $(u,v)$, and is often derived from correlation scores. While implementation-specific, this map provides a per-pixel measure of the trustworthiness of the SL depth.

Following~\cite{antensteiner2018review}, we estimate a fused depth map $\hat{d}(u,v)$   
from the observed depth map \(d_\text{obs}(u,v)\) and the gradient field \(\mathbf g_\text{obs}(u,v)\).  The latter is derived from PS normal map $\mathbf n_\text{obs}(u,v)$ using the relation in \eqref{eq:grad_from_normal}. 
The fusion is formulated as the following optimization problem:\vspace{-2mm}
\begin{align}
\hat{d}(u,v)& =
\arg \underset{d(u,v)}\min \bigg( \alpha\!\!\!\!\sum_{(u,v)\in\mathcal P} \!\!\! \kappa(u,v)
\bigl(d(u,v)-d_\text{obs}(u,v)\bigr)^2
\nonumber \\ & +\;
\beta\!\!\sum_{(u,v)\in\mathcal P} \!\!
\bigl\|\nabla d(u,v)-\mathbf g_\text{obs}(u,v)\bigr\|_2^2 \bigg)
\label{eq:ortho_optimization_problem}
,
\end{align} 
where the optimized depth map \(d(u,\!v)\) is simultaneously guided towards agreement with both the observed depth map and the observed gradient field.
The parameters \(\alpha,\beta\ge 0\) balance the contributions of depth consistency and gradient consistency terms, respectively.
Because both terms in~\eqref{eq:ortho_optimization_problem} are quadratic in the unknown depth values, the objective is convex, and the optimization
problem can be solved efficiently.

In the regions of depth map, where no depth information is available, we want to perform inpainting based on the available surface normals. 
This is achieved by weighting the depth terms $\bigl(d(u,v)-d_\text{obs}(u,v)\bigr)^2$ by \(\kappa(u,v)\). Where \(\kappa(u,v)=0\), the observed depth values are completely ignored, and the reconstructed depth relies entirely on the gradient (and hence surface normal) information.

Alternative formulations that employ different objective functions for depth–normals fusion are reviewed in~\cite{antensteiner2018review}.
A particularly effective variant is the total generalized variation (TGV) method presented in \cite{antensteiner2018variational,antensteiner2018review,chambolle2011first}.
Relative to \eqref{eq:ortho_optimization_problem}, the TGV approach introduces an auxiliary gradient field $\mathbf q(u,v)$.
This field is simultaneously driven towards the gradient of the sought depth map 
$\nabla d(u,v)$ and the gradient $\mathbf g_\text{obs}(u,v)$ obtained from observed normals 
via~\eqref{eq:grad_from_normal}.
By penalizing not only the discrepancy between these quantities but also the spatial variation of 
$\mathbf q(u,v)$, TGV achieves depth smoothing effects while suppressing the staircasing artefacts characteristic of methods with first-order regularizers.
Using compact notation that omits the pixel coordinates 
$(u,v)$, the resulting optimization problem can be written as \vspace{-1mm}
\begin{align}
\begin{bmatrix}
\hat{d}(u,\!v) \\
\hat{\mathbf{q}}(u,\!v)
\end{bmatrix}
&=
\arg \! \underset{ \substack{d(u,v)\\ \mathbf{q}(u,v)}}\min \bigg( \alpha\!\!\!\!\sum_{(u,v)\in\mathcal P} \!\!\!\!\! \kappa
\bigl(d-d_\text{obs}\bigr)^2 + \lambda_0\!\!\!\!\sum_{(u,v)\in\mathcal P} \!\!\! \bigl\| \nabla \mathbf{q} \bigr\|_2
\nonumber \\ & +\;
\lambda_1\!\!\!\!\!\sum_{(u,v)\in\mathcal P} \!\!\!\!\bigl\| \nabla {d} - \mathbf{q} \bigr\|_2 +
\beta\!\!\sum_{(u,v)\in\mathcal P} \!\!\!\!
\bigl\|\mathbf{q}-\mathbf g_\text{obs}\bigr\|_2^2 \bigg)
,
\label{eq:ortho_optimization_problem_tgv}
\end{align} \nopagebreak
 \nopagebreak where \(\alpha,\beta\ge 0\) retain their roles from~\eqref{eq:ortho_optimization_problem}, and \(\lambda_0,\lambda_1\ge 0\) balance the first- and second-order TGV regularization terms.

\vspace{0mm}
\section{Proposed perspective fusion}
\label{sec:proposed_method}

Under perspective projection the 3D point $\mathbf{s}$ corresponding to the image point $\mathbf{u}$ is  
\vspace{-2mm}
\begin{equation}
\mathbf{s}_p(u,v)=
\begin{bmatrix}
\dfrac{d(u,v)\,\bigl(u-c_u\bigr)}{f}\\[6pt]
\dfrac{d(u,v)\,\bigl(v-c_v\bigr)}{f}\\[6pt]
d(u,v)
\end{bmatrix},
\label{eq:perspective_point}
\end{equation}
where \(f\) is the focal length and \((c_u,c_v)\) is the principal point.
The two surface tangent vectors then are \vspace{-1.5mm}
\begin{equation}
\partial_u\mathbf s_p(u,v) =
\begin{bmatrix}
\frac{1}{f}((u-c_u)\partial_u d(u,v) + d(u,v)) \\[2pt]
\frac{1}{f}(v-c_v)\partial_u d(u,v) \\[2pt]
\partial_u d(u,v)
\end{bmatrix}
\end{equation}
and \vspace{-1mm}
\begin{equation}
\partial_v\mathbf s_p(u,v) =
\begin{bmatrix}
\frac{1}{f}(u-c_u)\partial_v d(u,v) \\[2pt]
\frac{1}{f}((v-c_v)\partial_v d(u,v) + d(u,v)) \\[2pt]
\partial_v d(u,v)
\end{bmatrix}.
\end{equation}
Computing the cross product $\mathbf{n}_p(u,v) \propto \partial_u\mathbf{s}_p(u,v) \times \partial_v\mathbf{s}_p(u,v)$ (cf. \eqref{eq:normal_def}) (we omit the resulting expression for brevity), one can see that the simple relationship \eqref{eq:grad_from_normal} between the normal vector $\mathbf{n}(u,v)$ and the gradient $\mathbf{g}(u,v)$ of the depth map does not hold under the perspective projection.

To regain the convenient gradient-based formulation of
fusion from 
Section \emph{\nameref{sec:ortho_fusion}},
%Section \ref{sec:ortho_fusion},
we adopt the
\emph{log-depth substitution} introduced for normal integration in
\cite{queau2018normal,cao2022bilateral} and extend it to the joint
fusion of depth and normals.
In \cite{queau2018normal,cao2022bilateral}, it is shown that after performing a log-depth substitution \vspace{-1mm}
\begin{equation}
l(u,v)\;=\;\ln d(u,v)\quad (d>0), %\vspace{-1mm}
\label{eq:logdepth_def}
\end{equation}
one can derive the following relationship between the normals $\mathbf n (u,v)$
and the gradient field of the log-depth $l(u,v)$: \vspace{-1mm}
\begin{equation}
\begin{aligned}
\nabla l(u,v)\;=\;
\tilde{\mathbf g}(u,v)
\;=\;
\begin{bmatrix}
\partial_u l(u,v) \\ \partial_v l(u,v)
\end{bmatrix} \\
\;=\;
-\frac{1}{\,(u-c_u)\,n_x + (v-c_v)\,n_y + f\,n_z\,}
\begin{bmatrix}
n_x\\[2pt] n_y
\end{bmatrix}.
\label{eq:persp_grad_from_normals}
\end{aligned}
\end{equation}
We refer the reader to  \cite{queau2018normal,cao2022bilateral} for the mathematical derivation.
Equation \eqref{eq:persp_grad_from_normals} plays the same role for fusion under 
perspective projection as \eqref{eq:grad_from_normal} plays in the
orthographic case, i.e., it establishes a relationship between the observed surface normal vectors  
and the gradient of the observed log-depth map.
Leveraging this relationship, we reformulate the optimization problem~\eqref{eq:ortho_optimization_problem} to operate on the log-depth map
\(l(u,\!v)\) and the transformed normal-derived gradient  field
\(\tilde{\mathbf g}(u,\!v)\). 
The result is a convex optimization problem—analogous to the orthographic case \eqref{eq:ortho_optimization_problem}—and its solution yields the fused log-depth map
\(\hat{l}(u,\,v)\): \vspace{-1.5mm}
\begin{align}
\hat{l}(u,v) & = \arg \underset{l(u,v)}\min \bigg(
\alpha\!\!\!\!\sum_{(u,v)\in\mathcal P} \!\!\!\! \kappa(u,v)
\bigl(l(u,v)-\ln d_\text{obs}(u,v)\bigr)^2 \nonumber
\\[0mm] & +\; 
\beta\!\!\sum_{(u,v)\in\mathcal P}
\bigl\|\nabla l(u,v)-\tilde{\mathbf g}_\text{obs}(u,v)\bigr\|_2^2
\bigg),
\label{eq:persp_grad_energy}
\end{align}
where \(\tilde{\mathbf g}_\text{obs}(u,\!v)\) is obtained from
\(\mathbf n_\text{obs}(u,\!v)\) via \eqref{eq:persp_grad_from_normals}.
The confidence map \(\kappa(u,v)\in[0,1]\) plays exactly the same role as
in the orthographic case, \(\alpha,\beta\ge 0\) balance the two
terms, and the objective remains convex in the
unknown log-depth \(l(u,v)\).
Pixels with \(\kappa(u,v)=0\) naturally fall back to pure normal integration,
providing principled \emph{perspective-aware inpainting} of regions where the depth sensing
fails. 
After optimization, the fused metric depth is recovered by simple
exponentiation: %\vspace{-2mm}
\begin{equation}
\hat{d}(u,v)=\exp \bigl(\hat{l}(u,v)\bigr). \vspace{-0.25mm}
\label{eq:exponentiation}
\end{equation}

This approach, summarized in Algorithm \ref{algo:perspective-fusion}, transforms perspective fusion into
the same
problem as 
explored in Section \emph{\nameref{sec:ortho_fusion}}.
%\ref{sec:ortho_fusion}.
Only the inputs change:
depth values are log transformed, and normals are converted to the
log-depth
gradient field
\(\tilde{\mathbf g}_\text{obs}(u,\!v)\) using \eqref{eq:persp_grad_from_normals}.
Existing gradient-based orthographic fusion methods carry over unchanged, while
full perspective consistency is retained. 
Using Algorithm \ref{algo:perspective-fusion}, we can easily extend the TGV-based fusion \eqref{eq:ortho_optimization_problem_tgv} to handle perspective projection. Note that by performing the log-depth substitution, the auxiliary gradient $\mathbf q(u,v)$ is interpreted as the gradient of the log-depth map (cf. \eqref{eq:persp_grad_from_normals}) and is denoted as $\mathbf{p}(u,v)$.
Performing the necessary substitutions in  \eqref{eq:ortho_optimization_problem_tgv}, 
the resulting optimization problem can be written as \vspace{-1.5mm}
\begin{align}
\begin{bmatrix}
\hat{l}(u,\!v) \\
\hat{\mathbf{p}}(u,\!v)
\end{bmatrix}
&=
\arg \! \underset{ \substack{l(u,v)\\ \mathbf{p}(u,v)}}\min \bigg( \alpha\!\!\!\!\sum_{(u,v)\in\mathcal P} \!\!\!\!\! \kappa
\bigl(l-\ln d_\text{obs}\bigr)^2 + \lambda_0\!\!\!\!\sum_{(u,v)\in\mathcal P} \!\!\! \bigl\| \nabla \mathbf{p} \bigr\|_2
\nonumber \\ & +\;
\lambda_1\!\!\!\!\!\sum_{(u,v)\in\mathcal P} \!\!\!\!\bigl\| \nabla {l} - \mathbf{p} \bigr\|_2 +
\beta\!\!\sum_{(u,v)\in\mathcal P} \!\!\!\!
\bigl\|\mathbf{p}-\tilde{\mathbf{g}}_\text{obs}\bigr\|_2^2 \bigg)
. \vspace{-1mm}
\label{eq:persp_optimization_problem_tgv}
\end{align}
This perspective-aware method retains the smoothing and staircasing-avoiding qualities of orthographic TGV fusion.

\vspace{-1mm}
\begin{algorithm}[h]
  \caption{Perspective-aware depth–normals fusion}\vspace{1mm}
\label{algo:perspective-fusion}
       
    \textbf{Inputs:} Observed depth map $d_\text{obs}(u,\!v)$, observed surface normal map $\mathbf{n}_\text{obs}(u,\!v)$
\vspace{1mm}
\begin{algorithmic}[1]    
    \State Transform depth $d_\text{obs}(u,\!v)$ to log-depth $\ln d_\text{obs}(u,\!v)$.
    \State Convert normals $\mathbf{n}_\text{obs}(u,\!v)$ to log-depth gradients $\tilde{\mathbf{g}}_\text{obs}(u,\!v)$ using \eqref{eq:persp_grad_from_normals}.
    \State Perform any gradient-based orthographic depth-normals fusion of $\ln d_\text{obs}(u,\!v)$ and $\tilde{\mathbf{g}}_\text{obs}(u,\!v)$ to obtain the fused log-depth $\hat l(u,\!v)$.

    \State Transform $\hat l(u,\!v)$ back to metric depth $\hat d(u,\!v)$ using  \eqref{eq:exponentiation}.

  \end{algorithmic}
\vspace{1mm}
\textbf{Output:} Fused depth map $\hat d(u,\!v)$
\end{algorithm}

\vspace{-4mm}
\section{Performance evaluation}
\label{sec:application}

\begin{figure}[htbp]
  \vspace{-2mm}
  \centering

  \subfloat[\tiny Reference depth]{\includegraphics[clip,width=0.24\linewidth]{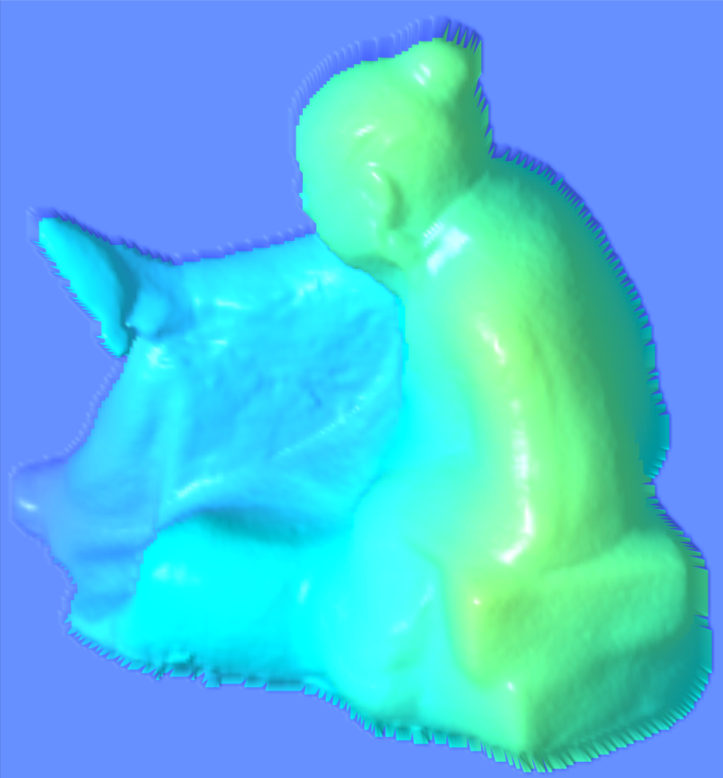}\label{fig:subfig_depth_GT}}\hfil
  \subfloat[\tiny Perturbed depth]{\includegraphics[clip,width=0.24\linewidth]{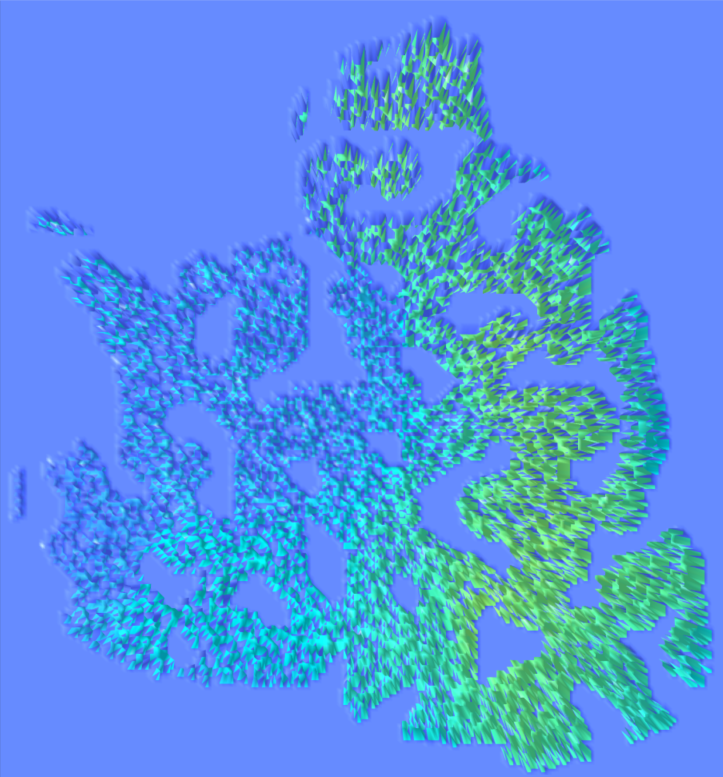}\label{fig:subfig_depth_holes}}\hfil
  \subfloat[\tiny Confidence]{\includegraphics[clip,width=0.24\linewidth]{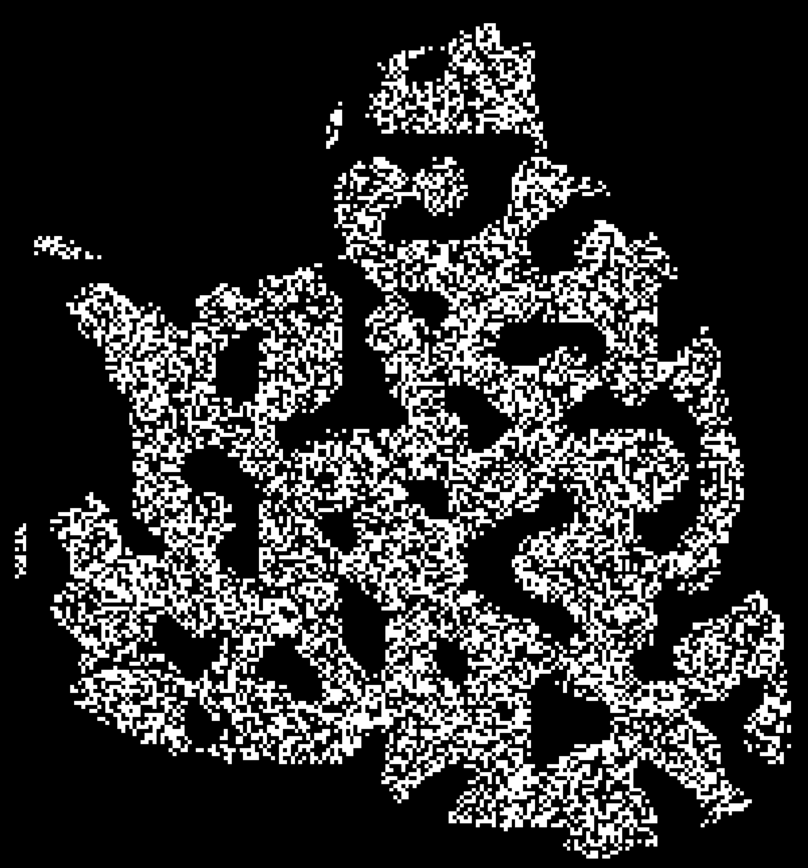}\label{fig:conf}}\hfil
  \subfloat[\tiny Normals]{\includegraphics[clip,width=0.24\linewidth]{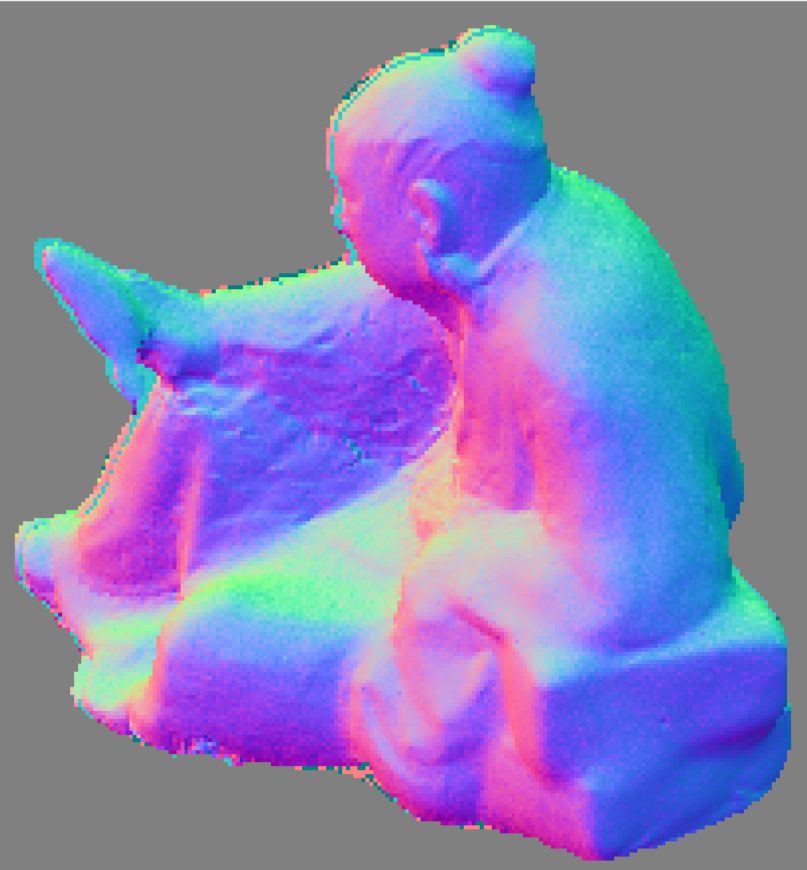}\label{fig:subfig_normals}}\\[0.4ex]
  \vspace{-3.0mm}
  \subfloat[\tiny "Ortho" depth]{\includegraphics[clip,width=0.24\linewidth]{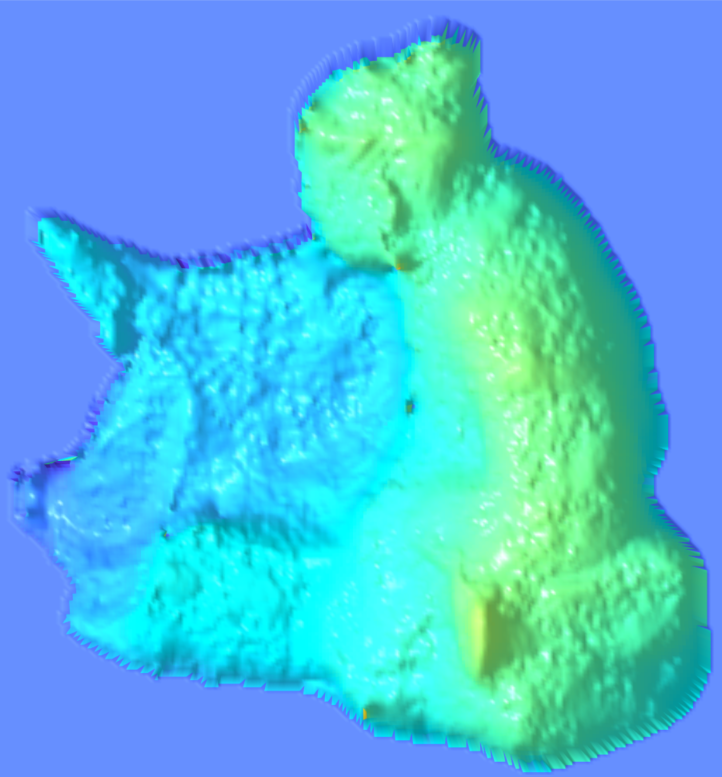}\label{fig:subfig_depth_ortho}}\hfil
  \subfloat[\tiny "Naive" depth]{\includegraphics[clip,width=0.241\linewidth]{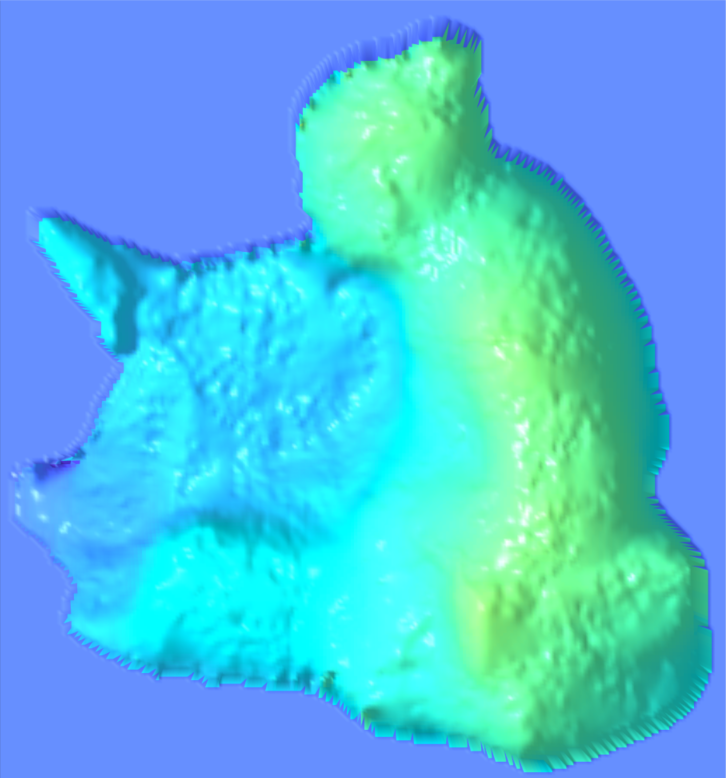}\label{fig:subfig_depth_naive}}\hfil
  \subfloat[\tiny "PG" depth]{\includegraphics[clip,width=0.24\linewidth]{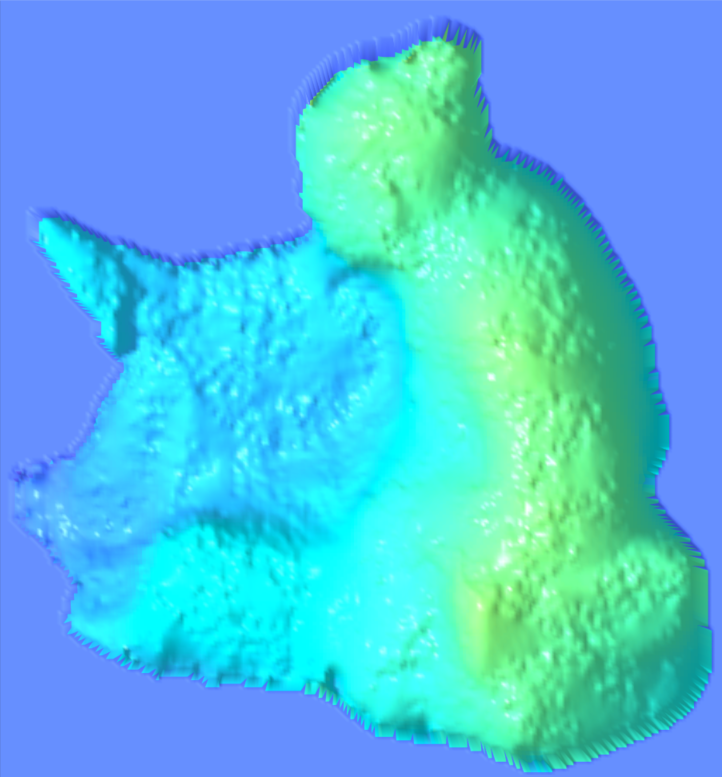}\label{fig:subfig_depth_grad}}\hfil
  \subfloat[\tiny "PTGV" depth]{\includegraphics[clip,width=0.241\linewidth]{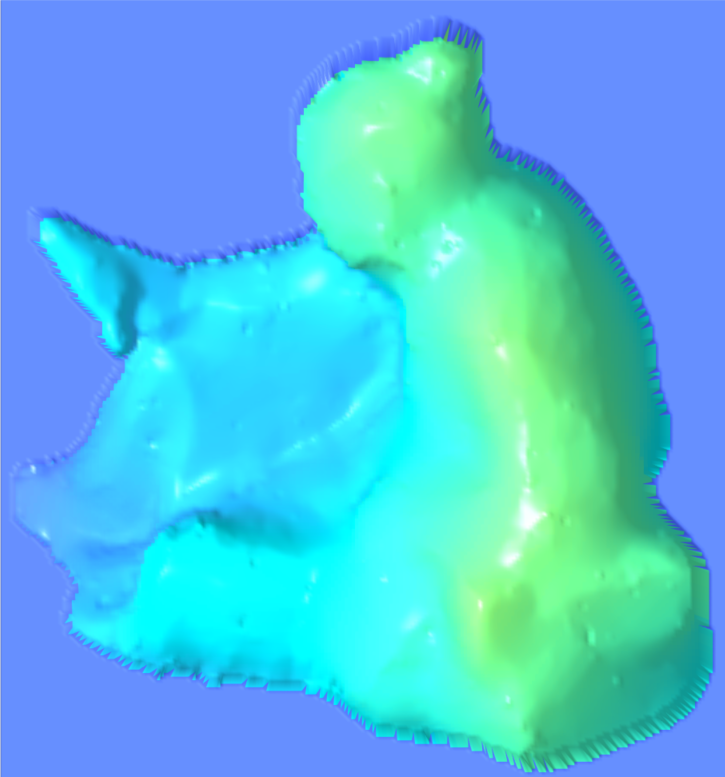}\label{fig:subfig_depth_tgv}} \vspace{-2mm}

  \caption{
(a) Ground truth depth data for the \emph{reading} object from the DiLiGenT-MV data set. (b) Lower-resolution depth data perturbed with noise and missing depth information. (c) Confidence map $\kappa$. (d) RGB color-coded surface normal map perturbed with noise.
(e–h) Surface reconstructions using \emph{Ortho} (e), \emph{Naive} (f), \emph{PG} (g), and \emph{PTGV} (h) fusion methods applied to (b) and (d).}
  \label{fig:depth_and_normals}
  \vspace{-6mm}
\end{figure}

We evaluate our perspective-aware fusion on the DiLiGenT-MV dataset \cite{Li2020DiLiGenT-MV,DiLiGenT-MV-Website}, which contains 3D meshes and surface normal maps for five objects (\emph{bear}, \emph{buddha}, \emph{cow}, \emph{pot2}, and \emph{reading}). For our experiments, we use the meshes and normals estimated by \cite{Park2017MVPS} that are part of the data set. We only consider the first view direction, for which we convert each 3D mesh into a perspective depth map using the provided camera intrinsics and extrinsics. The obtained depth map represents the ground truth that we use for performance evaluation. For the \emph{reading} object, we visualize it in Fig. \ref{fig:subfig_depth_GT}. To emulate a realistic SL depth map, we down-sample it by randomly discarding every second pixel, reflecting the lower resolution typical for SL sensors \cite{zanuttigh2016operating,zhang2012edge}.
We further introduce missing-depth regions \cite{wang2016filling,niu2025novel,traxler2021experimental,yu2013shading} to model SL sensor imperfections (e.g., shadowing and SL pattern reflections and absorption, leading to failures in feature extraction, correspondence, and triangulation) by masking pixels with a thresholded Perlin noise \cite{Perlin1985} field.
Pixels with Perlin noise values above a threshold are marked missing, yielding gaps in approximately 25\% of the object area.
Although this gap percentage is more severe than that of typical SL scanners \cite{traxler2021experimental}, we adopt it to stress-test our depth-normals fusion. 
Zero-mean Gaussian noise with $\sigma\!=\!1.0\,\text{mm}$ is then added to the depth values. The resulting SL depth map (shown in Fig. \ref{fig:subfig_depth_holes}) serves as input to the fusion algorithms. Pixels with missing depth information are indicated in the confidence map by \(\kappa(u,v)\!=\!0\) (Fig. \ref{fig:conf}). The normal map is perturbed with zero-mean Gaussian noise of $\sigma\!=\!0.1$, but remains at full resolution and without gaps (see Fig. \ref{fig:subfig_normals}), reflecting the characteristics of a PS system, where normals can always be estimated for each pixel, since no feature extraction, correspondence, and triangulation is required.

To evaluate performance, we compute the root mean squared error (RMSE) between the ground truth depth map $d_{\text{GT}}(u,v)$ and the reconstructed (fused) depth map $\hat{d}(u,v)$ as
$\text{RMSE} = \sqrt{1/|\mathcal{P}| %\frac{1}{|\mathcal{P}|} 
\sum_{(u,v)\in\mathcal{P}} ( \hat{d}(u,v) - d_{\text{GT}}(u,v) )^2}.$
We also compute the mean angular error (MAE) between the estimated surface normals $\hat{\mathbf{n}}(u,v)$ (computed from fused depth $\hat{d}(u,v)$) and the ground truth normals $\mathbf{n}_{\text{GT}}(u,v)$ as
$\text{MAE} = 1/|\mathcal{P}| %\frac{1}{|\mathcal{P}|} 
\sum_{(u,v)\in\mathcal{P}} \arccos \left( \hat{\mathbf{n}}(u,v) \cdot \mathbf{n}_{\text{GT}}(u,v) \right).$

We compare pure orthographic fusion \eqref{eq:ortho_optimization_problem} directly applied to perspective depth maps (denoted \emph{Ortho})  with several perspective-aware fusion approaches. The first perspective-aware approach transforms the perspective depth map to an orthographic one using the reprojection approach described in Section \emph{\nameref{sec:alt_fusion}}
%~\ref{sec:alt_fusion} 
and afterwards uses \eqref{eq:ortho_optimization_problem}
(denoted \emph{Naive}). Another method is the approach from \cite{nehab2005efficiently} extended by us to handle missing depth data (denoted \emph{Nehab}\footnote{We use the MATLAB code developed for \cite{yu2019depth}, available online at: https://github.com/waps101/MergePositionNormals.}). We also include the fusion method proposed in \cite{cao2022code} (denoted \emph{BiNI}). Our own perspective-aware gradient-based fusion using \eqref{eq:persp_grad_energy}, is also evaluated (denoted \emph{PG}), as well as our perspective-aware TGV fusion method from \eqref{eq:persp_optimization_problem_tgv} (denoted \emph{PTGV}).

In Fig. \ref{fig:subfig_depth_ortho}-\ref{fig:subfig_depth_tgv}, we show the reconstruction results using the \emph{Ortho} (Fig. \ref{fig:subfig_depth_ortho}), \emph{Naive} (Fig. \ref{fig:subfig_depth_naive}), \emph{PG} (Fig. \ref{fig:subfig_depth_grad}) and  \emph{PTGV} (Fig. \ref{fig:subfig_depth_tgv}) fusion methods using the SL depth map from Fig. \ref{fig:subfig_depth_holes} and the PS surface normal map from Fig. \ref{fig:subfig_normals} as inputs.

In Table~\ref{tab:metrics_comparison_with_holes}, we present the RMSE and MAE of the fused depth maps for each of the $5$ objects from the DiLiGenT-MV data set. We also present the average RMSE and MAE values computed over all data set objects.

To obtain these results, the following parameters are used. \emph{Ortho}, \emph{Naive}, \emph{PG}, and \emph{PTGV} all use $\alpha=1$, $\beta=1$, and (required only by \emph{PTGV}) $\lambda_0=\lambda_1=0.001$. \emph{Nehab} uses $\lambda = 0.1$, and finally  \emph{BiNI} uses $\lambda_1 = 10^7$.

We see that \emph{PTGV} achieves the lowest average RMSE and MAE values. This is expected as TGV has the best noise suppression properties.
Our log-depth approach makes it directly applicable to perspective maps in a straightforward fashion.
Furthermore, we can see that \emph{PG} improves over the \emph{Ortho} and \emph{Naive} methods in RMSE. Applying \eqref{eq:ortho_optimization_problem} directly to perspective depth maps (\emph{Ortho}) performs worst, highlighting the need to account for perspective projection. Reprojecting to orthographic maps first and then applying \eqref{eq:ortho_optimization_problem} (\emph{Naive}) recovers part of the performance degradation but still lags behind methods that treat perspective natively, such as \emph{PG} which uses \eqref{eq:persp_grad_energy}. Overall, these results indicate that considering perspective is necessary; full perspective-aware fusion is recommended when fusing perspective SL depth maps with PS normals.

\vspace{-2.0mm}
\begin{table}[h]
\vspace{-1.0mm}
\centering
\setlength{\tabcolsep}{4pt} 
\renewcommand{\arraystretch}{0.75}
\setlength{\doublerulesep}{0.05ex}   % gap between the two lines
\hspace{5mm}
\caption{Evaluation of fusion methods on SL depth and PS normal \\data derived from the DiLiGenT-MV data set. RMSE is in [mm] and MAE in [rad]. }
\label{tab:metrics_comparison_with_holes}
\vspace{0.0mm}
\begin{tabular}{llcccccc}
\hline
\vspace{-0.5mm}    & & Ortho & Naive & Nehab & BiNI & PG & PTGV \\
\hline \hline
bear \vspace{-1mm}  & RMSE $\!\downarrow$     &    1.506 & 0.728 & 0.392 & 0.463 & 0.634 & \bf{0.353}      \\ 

\vspace{-0.5mm} & MAE $\!\downarrow$     &     0.431 & 0.259 & 0.530 & 0.398 & 0.371 & \bf{0.129}     \\
\hline
buddha \vspace{-1mm} & RMSE $\!\downarrow$     &    2.245 & 1.678 & 2.731 & 1.847 & 1.753 & \bf{1.056}    \\

\vspace{-0.5mm} & MAE $\!\downarrow$     &     0.500 & 0.342 & 0.672 & 0.463 & 0.446 & \bf{0.279}     \\
\hline
cow \vspace{-1mm} & RMSE $\!\downarrow$     &    1.656 & 1.309 & 0.596 & 0.663 & 1.229 & \bf{0.653}     \\

\vspace{-0.5mm} & MAE $\!\downarrow$     &      0.468 & 0.307 & 0.572 & 0.417 & 0.412 & \bf{0.153}    \\
\hline
pot2 \vspace{-1mm} & RMSE $\!\downarrow$     &   2.746 & 1.526 & \bf{0.467} & 0.530 & 0.781 & 0.610     \\

\vspace{-0.5mm} & MAE $\!\downarrow$     &      0.445 & 0.279 & 0.566 & 0.416 & 0.373 & \bf{0.177}  \\
\hline
reading \vspace{-1mm} & RMSE $\!\downarrow$     &    2.616 & 2.002 & 2.313 & \bf{1.068} & 1.705 & 1.140 \\

\vspace{-0.5mm} & MAE $\!\downarrow$     &       0.489 & 0.336 & 0.635 & 0.451 & 0.414 & \bf{0.238}  \\
\hline \hline
Average \vspace{-1mm} & RMSE $\!\downarrow$     &  2.154 & 1.449 & 1.300 & 0.914 & 1.220 & \bf{0.762}       \\

\vspace{-0.5mm} & MAE $\!\downarrow$     &      0.467 & 0.305 & 0.586 & 0.429 & 0.403 & \bf{0.195} \\
\hline
\end{tabular}
\vspace{-1mm}
\end{table}

\vspace{-5mm}
\section{Conclusion}

We propose a novel perspective-aware fusion method for combining depth and surface normal maps. The method employs a log-depth substitution, which transforms the perspective fusion problem into an orthographic one. This transformation enables us to extend any suitable orthographic fusion approach to handle perspective data.
Additionally, we show that our method naturally handles missing depth information and can perform perspective-aware inpainting using surface normals.
Our evaluation on the DiLiGenT-MV data set demonstrates the effectiveness of the proposed approach and the importance of perspective awareness in depth-normals fusion.

\balance

\bibliography{references}

\begin{thebibliography}{10}
\ifx\bibinfo\relax\providecommand{\bibinfo}[2]{#2}\fi
\makeatletter
\ifx\xfnm\@undefined\gdef\xfnm[#1]{#1}\fi
\ifx\xsnm\@undefined\gdef\xsnm[#1]{#1}\fi
\ifx\plxcitation\@undefined\def\plxcitation#1#2#3#4#5{}\fi
\ifx\endplxcitation\@undefined\def\endplxcitation{}\fi
\makeatother

\bibitem{zanuttigh2016operating}
\plxcitation{}{zanuttigh2016operating}{}{}{article}
\bibinfo{author}{\xsnm[Zanuttigh]\xfnm[, P.]}, et~al.:
  \bibinfo{title}{Operating principles of structured light depth cameras}.
\newblock \bibinfo{journal}{Time-of-Flight and Structured Light Depth Cameras:
  Technology and Applications} pp.~ \bibinfo{pages}{43--79}.
  (\bibinfo{year}{2016})
\endplxcitation

\bibitem{woodham1980photometric}
\plxcitation{}{woodham1980photometric}{}{}{article}
\bibinfo{author}{\xsnm[Woodham]\xfnm[, R.J.]}: \bibinfo{title}{Photometric
  method for determining surface orientation from multiple images}.
\newblock \bibinfo{journal}{Optical Engineering}
  \bibinfo{volume}{19}(\bibinfo{number}{1}), \bibinfo{pages}{139--144}
  (\bibinfo{year}{1980})
\endplxcitation

\bibitem{ackermann2015survey}
\plxcitation{}{ackermann2015survey}{}{}{article}
\bibinfo{author}{\xsnm[Ackermann]\xfnm[, J.]},
  \bibinfo{author}{\xsnm[Goesele]\xfnm[, M.]}, et~al.: \bibinfo{title}{A survey
  of photometric stereo techniques}.
\newblock \bibinfo{journal}{Foundations and Trends in Computer Graphics and
  Vision} \bibinfo{volume}{9}(\bibinfo{number}{3-4}), \bibinfo{pages}{149--254}
  (\bibinfo{year}{2015})
\endplxcitation

\bibitem{haque2014high}
\plxcitation{}{haque2014high}{}{}{inproceedings}
\bibinfo{author}{\xsnm[Haque]\xfnm[, M.]}, et~al.: \bibinfo{title}{High quality
  photometric reconstruction using a depth camera}.
\newblock In: \bibinfo{booktitle}{Proceedings of the IEEE Conference on
  Computer Vision and Pattern Recognition}, pp.~ \bibinfo{pages}{2275--2282}.
  (\bibinfo{year}{2014})
\endplxcitation

\bibitem{antensteiner2018review}
\plxcitation{}{antensteiner2018review}{}{}{article}
\bibinfo{author}{\xsnm[Antensteiner]\xfnm[, D.]},
  \bibinfo{author}{\xsnm[{\v{S}}tolc]\xfnm[, S.]},
  \bibinfo{author}{\xsnm[Pock]\xfnm[, T.]}: \bibinfo{title}{A review of depth
  and normal fusion algorithms}.
\newblock \bibinfo{journal}{Sensors} \bibinfo{volume}{18}(\bibinfo{number}{2}),
  \bibinfo{pages}{431} (\bibinfo{year}{2018})
\endplxcitation

\bibitem{antensteiner2017high}
\plxcitation{}{antensteiner2017high}{}{}{article}
\bibinfo{author}{\xsnm[Antensteiner]\xfnm[, D.]}, et~al.:
  \bibinfo{title}{High-precision 3{D} sensing with hybrid light field
  photometric stereo approach in multi-line scan framework}.
\newblock \bibinfo{journal}{Electronic Imaging} \bibinfo{volume}{29},
  \bibinfo{pages}{52--60} (\bibinfo{year}{2017})
\endplxcitation

\bibitem{antensteiner2018variational}
\plxcitation{}{antensteiner2018variational}{}{}{inproceedings}
\bibinfo{author}{\xsnm[Antensteiner]\xfnm[, D.]},
  \bibinfo{author}{\xsnm[{\v{S}}tolc]\xfnm[, S.]},
  \bibinfo{author}{\xsnm[Pock]\xfnm[, T.]}: \bibinfo{title}{Variational fusion
  of light field and photometric stereo for precise 3{D} sensing within a
  multi-line scan framework}.
\newblock In: \bibinfo{booktitle}{Proceedings of the 24th International
  Conference on Pattern Recognition}, pp.~ \bibinfo{pages}{1036--1042}.
  (\bibinfo{year}{2018})
\endplxcitation

\bibitem{wang2016filling}
\plxcitation{}{wang2016filling}{}{}{article}
\bibinfo{author}{\xsnm[Wang]\xfnm[, Z.]}, et~al.: \bibinfo{title}{Filling
  {K}inect depth holes via position-guided matrix completion}.
\newblock \bibinfo{journal}{Neurocomputing} \bibinfo{volume}{215},
  \bibinfo{pages}{48--52} (\bibinfo{year}{2016})
\endplxcitation

\bibitem{niu2025novel}
\plxcitation{}{niu2025novel}{}{}{article}
\bibinfo{author}{\xsnm[Niu]\xfnm[, Z.]}, et~al.: \bibinfo{title}{A novel
  approach to optimize key limitations of {A}zure {K}inect {DK} for efficient
  and precise leaf area measurement}.
\newblock \bibinfo{journal}{Agriculture}
  \bibinfo{volume}{15}(\bibinfo{number}{2}) (\bibinfo{year}{2025})
\endplxcitation

\bibitem{traxler2021experimental}
\plxcitation{}{traxler2021experimental}{}{}{article}
\bibinfo{author}{\xsnm[Traxler]\xfnm[, L.]}, et~al.:
  \bibinfo{title}{Experimental comparison of optical inline {3D} measurement
  and inspection systems}.
\newblock \bibinfo{journal}{IEEE Access} \bibinfo{volume}{9},
  \bibinfo{pages}{53952--53963} (\bibinfo{year}{2021})
\endplxcitation

\bibitem{yu2013shading}
\plxcitation{}{yu2013shading}{}{}{inproceedings}
\bibinfo{author}{\xsnm[Yu]\xfnm[, L.F.]}, et~al.: \bibinfo{title}{Shading-based
  shape refinement of {RGB-D} images}.
\newblock In: \bibinfo{booktitle}{Proceedings of the IEEE Conference on
  Computer Vision and Pattern Recognition}, pp.~ \bibinfo{pages}{1415--1422}.
  (\bibinfo{year}{2013})
\endplxcitation

\bibitem{queau2018normal}
\plxcitation{}{queau2018normal}{}{}{article}
\bibinfo{author}{\xsnm[Qu{\'e}au]\xfnm[, Y.]},
  \bibinfo{author}{\xsnm[Durou]\xfnm[, J.D.]},
  \bibinfo{author}{\xsnm[Aujol]\xfnm[, J.F.]}: \bibinfo{title}{Normal
  integration: {A} survey}.
\newblock \bibinfo{journal}{Journal of Mathematical Imaging and Vision}
  \bibinfo{volume}{60}, \bibinfo{pages}{576--593} (\bibinfo{year}{2018})
\endplxcitation

\bibitem{cao2022bilateral}
\plxcitation{}{cao2022bilateral}{}{}{inproceedings}
\bibinfo{author}{\xsnm[Cao]\xfnm[, X.]}, et~al.: \bibinfo{title}{Bilateral
  normal integration}.
\newblock In: \bibinfo{booktitle}{Proceedings of the European Conference on
  Computer Vision}, pp.~ \bibinfo{pages}{552--567}.  (\bibinfo{year}{2022})
\endplxcitation

\bibitem{kim2024discontinuity}
\plxcitation{}{kim2024discontinuity}{}{}{inproceedings}
\bibinfo{author}{\xsnm[Kim]\xfnm[, H.]}, \bibinfo{author}{\xsnm[Jung]\xfnm[,
  Y.]}, \bibinfo{author}{\xsnm[Lee]\xfnm[, S.]}:
  \bibinfo{title}{Discontinuity-preserving normal integration with auxiliary
  edges}.
\newblock In: \bibinfo{booktitle}{Proceedings of the IEEE Conference on
  Computer Vision and Pattern Recognition}, pp.~ \bibinfo{pages}{11915--11923}.
   (\bibinfo{year}{2024})
\endplxcitation

\bibitem{frankot1988method}
\plxcitation{}{frankot1988method}{}{}{article}
\bibinfo{author}{\xsnm[Frankot]\xfnm[, R.T.]},
  \bibinfo{author}{\xsnm[Chellappa]\xfnm[, R.]}: \bibinfo{title}{A method for
  enforcing integrability in shape from shading algorithms}.
\newblock \bibinfo{journal}{IEEE Transactions on Pattern Analysis and Machine
  Intelligence} \bibinfo{volume}{10}(\bibinfo{number}{4}),
  \bibinfo{pages}{439--451} (\bibinfo{year}{1988})
\endplxcitation

\bibitem{horn1986variational}
\plxcitation{}{horn1986variational}{}{}{article}
\bibinfo{author}{\xsnm[Horn]\xfnm[, B.K.]},
  \bibinfo{author}{\xsnm[Brooks]\xfnm[, M.J.]}: \bibinfo{title}{The variational
  approach to shape from shading}.
\newblock \bibinfo{journal}{Computer Vision, Graphics, and Image Processing}
  \bibinfo{volume}{33}(\bibinfo{number}{2}), \bibinfo{pages}{174--208}
  (\bibinfo{year}{1986})
\endplxcitation

\bibitem{cao2022code}
\bibinfo{author}{\xsnm[Cao]\xfnm[, X.]}: \bibinfo{title}{Bilateral normal
  integration}.
\newblock \bibinfo{howpublished}{GitHub repository}.
\newblock  \url{https://github.com/xucao-42/bilateral_normal_integration}
  (\bibinfo{year}{2022})
\endplxcitation

\bibitem{cao2024supernormal}
\plxcitation{}{cao2024supernormal}{}{}{inproceedings}
\bibinfo{author}{\xsnm[Cao]\xfnm[, X.]},
  \bibinfo{author}{\xsnm[Taketomi]\xfnm[, T.]}: \bibinfo{title}{Supernormal:
  {N}eural surface reconstruction via multi-view normal integration}.
\newblock In: \bibinfo{booktitle}{Proceedings of the IEEE/CVF Conference on
  Computer Vision and Pattern Recognition}, pp.~ \bibinfo{pages}{20581--20590}.
   (\bibinfo{year}{2024})
\endplxcitation

\bibitem{nehab2005efficiently}
\plxcitation{}{nehab2005efficiently}{}{}{article}
\bibinfo{author}{\xsnm[Nehab]\xfnm[, D.]}, et~al.: \bibinfo{title}{Efficiently
  combining positions and normals for precise 3{D} geometry}.
\newblock \bibinfo{journal}{ACM Transactions on Graphics}
  \bibinfo{volume}{24}(\bibinfo{number}{3}), \bibinfo{pages}{536--543}
  (\bibinfo{year}{2005})
\endplxcitation

\bibitem{zhang2012edge}
\plxcitation{}{zhang2012edge}{}{}{inproceedings}
\bibinfo{author}{\xsnm[Zhang]\xfnm[, Q.]}, et~al.:
  \bibinfo{title}{Edge-preserving photometric stereo via depth fusion}.
\newblock In: \bibinfo{booktitle}{Proceedings of the IEEE Conference on
  Computer Vision and Pattern Recognition}, pp.~ \bibinfo{pages}{2472--2479}.
  (\bibinfo{year}{2012})
\endplxcitation

\bibitem{yu2019depth}
\plxcitation{}{yu2019depth}{}{}{inproceedings}
\bibinfo{author}{\xsnm[Yu]\xfnm[, Y.]}, \bibinfo{author}{\xsnm[Smith]\xfnm[,
  W.A.]}: \bibinfo{title}{Depth estimation meets inverse rendering for single
  image novel view synthesis}.
\newblock In: \bibinfo{booktitle}{Proceedings of the 16th ACM SIGGRAPH European
  Conference on Visual Media Production}, pp.~ \bibinfo{pages}{1--7}.
  (\bibinfo{year}{2019})
\endplxcitation

\bibitem{chambolle2011first}
\plxcitation{}{chambolle2011first}{}{}{article}
\bibinfo{author}{\xsnm[Chambolle]\xfnm[, A.]},
  \bibinfo{author}{\xsnm[Pock]\xfnm[, T.]}: \bibinfo{title}{A first-order
  primal-dual algorithm for convex problems with applications to imaging}.
\newblock \bibinfo{journal}{Journal of Mathematical Imaging and Vision}
  \bibinfo{volume}{40}, \bibinfo{pages}{120--145} (\bibinfo{year}{2011})
\endplxcitation

\bibitem{Li2020DiLiGenT-MV}
\plxcitation{}{Li2020DiLiGenT-MV}{}{}{article}
\bibinfo{author}{\xsnm[Li]\xfnm[, M.]}, et~al.: \bibinfo{title}{Multi-view
  photometric stereo: A robust solution and benchmark dataset for spatially
  varying isotropic materials}.
\newblock \bibinfo{journal}{IEEE Transactions on Image Processing}
  \bibinfo{volume}{29}(\bibinfo{number}{1}), \bibinfo{pages}{4159--4173}
  (\bibinfo{year}{2020})
\endplxcitation

\bibitem{DiLiGenT-MV-Website}
\bibinfo{author}{\xsnm[Li]\xfnm[, M.]}, et~al.: \bibinfo{title}{{DiLiGenT-MV
  Dataset}}.
\newblock \bibinfo{howpublished}{Website}.
\newblock  \url{https://sites.google.com/site/photometricstereodata/mv}
  (\bibinfo{year}{2020})
\endplxcitation

\bibitem{Park2017MVPS}
\plxcitation{}{Park2017MVPS}{}{}{article}
\bibinfo{author}{\xsnm[Park]\xfnm[, J.]}, et~al.: \bibinfo{title}{Robust
  multiview photometric stereo using planar mesh parameterization}.
\newblock \bibinfo{journal}{IEEE Transactions on Pattern Analysis and Machine
  Intelligence} \bibinfo{volume}{39}(\bibinfo{number}{8}),
  \bibinfo{pages}{1591--1604} (\bibinfo{year}{2017})
\endplxcitation

\bibitem{Perlin1985}
\plxcitation{}{Perlin1985}{}{}{inproceedings}
\bibinfo{author}{\xsnm[Perlin]\xfnm[, K.]}: \bibinfo{title}{An image
  synthesizer}.
\newblock In: \bibinfo{booktitle}{Proceedings of the 12th Annual Conference on
  Computer Graphics and Interactive Techniques (SIGGRAPH '85)}, pp.~
  \bibinfo{pages}{287--296}. ACM (\bibinfo{year}{1985})
\endplxcitation

\end{thebibliography}
\bibliographystyle{iet}

\end{document}